# Morphological Tagging and Lemmatization of Albanian: A Manually Annotated Corpus and Neural Models

Nelda Kote[1], Marenglen Biba[2], Jenna Kanerva[3], Samuel Rönnqvist[3] and Filip Ginter[3]

[1]Faculty of Information Technology, Polytechnic University of Tirana, Albania
[2]Faculty of Information Technology, New York University of Tirana, Albania
[3]TurkuNLP Group, University of Turku, Finland
nkote@fti.edu.al, marenglenbiba@unyt.edu.al, {jmnybl, saanro, figint}@utu.fi

**Abstract**
In this paper, we present the first publicly available part-of-speech and morphologically tagged corpus for the Albanian language, as well as a neural morphological tagger and lemmatizer trained on it. There is currently a lack of available NLP resources for Albanian, and its complex grammar and morphology present challenges to their development. We have created an Albanian part-of-speech corpus based on the Universal Dependencies schema for morphological annotation, containing about 118,000 tokens of naturally occuring text collected from different text sources, with an addition of 67,000 tokens of artificially created simple sentences used only in training. On this corpus, we subsequently train and evaluate segmentation, morphological tagging and lemmatization models, using the Turku Neural Parser Pipeline. On the held-out evaluation set, the model achieves 92.74% accuracy on part-of-speech tagging, 85.31% on morphological tagging, and 89.95% on lemmatization. The manually annotated corpus, as well as the trained models are available under an open license.

## 1. Introduction

The Albanian language is an Indo-European language spoken by around 7 million native speakers mostly in the area of Albania, Kosovo, and other regions of the Balkans. It creates an independent branch of the Indo-European language family. The language has been influenced by different languages such as Greek, Turkish, Slavic and Latin. The Albanian language is a unique and interesting language to study, with a complex grammar and inflectional paradigm that make the process of morphological tagging and lemmatization particularly challenging. It is a low-resource language that, to our knowledge, lacks openly available morphologically annotated corpora and tools for lemmatization, morphological analysis and part-of-speech tagging.

The aim of our work is to create an openly available corpus of manually annotated part-of-speech tags, morphological features and lemmas. We have collected a corpus of 118,000 tokens from different text sources, including sentences from a fiction book, grammar books, web crawls and Wikipedia. Additionally, we include a 67,000 token sample of artificially created simple sentences. The full corpus is annotated according to the Universal Dependencies (UD) (Nivre et al., 2016) tagset. Furthermore, this annotated corpus is used to train and evaluate neural models for segmentation, tagging and lemmatization using the Turku Neural Parser Pipeline (Kanerva et al., 2018). The annotated corpus as well as all trained models are freely available online.[1]

The paper is structured as follows. In Section 2, we discuss the previous research done in this field. In Section 3, we analyse the morphology of the Albanian language. In Section 4, we present the corpus and the Universal Dependencies tagset used in annotation. In Section 5, we describe the Turku Neural Parser Pipeline, and in Section 6 we show the experimental results of the trained models. In Section 7, we conclude our work.

## 2. Related Work

There are several previous attempts to develop NLP tools for Albanian. Below we will discuss the most relevant related work in detail. Most of the existing systems are rule or dictionary-based and not available online for NLP purposes. While there exist prior manually tagged corpora, none of them are publicly available.

### 2.1 Part-of-speech and Morphological Tagging

In Trommer and Kallulli (2004), the authors present a simple morphological analyzer for standard Albanian language. The morphological analyzer implements 340 morphological rules that covers the main inflection types of the Albanian language. The tagset contains 17 labels conforming to the EAGLE guidelines standard adapted to the description of the Albanian language. The morphological analyzer is evaluated using a small corpus of 1,000 tokens (words) achieving oracle precision of 97% (i.e. not taking into account the need for subsequent disambiguation), and recall of 92–95%.

Piton et al. (2007) and Piton and Lagji (2007) present an electronic dictionary using Finite State Transducers with NooJ's graphs for text in the Albanian language. The focus of the analysis is not the text but the word. An Albanian-French dictionary containing 4,951 words is used to build the flex forms dictionary. The number of tags is not specified in the paper, but at least the tags noun, verb (active and non-active), adjective, preposition, adverbs, interjection, cardinal and ordinal numbers are reported on. A special focus is given to augmented words, but the proposed solution is not experimentally evaluated.

Hasanaj and Biba (2011) and Salavaçi and Biba (2012) present the work to develop statistical part-of-speech tagging models for Albanian language using the OpenNLP library. A maximum entropy and a neural network model are trained and evaluated using different size tagsets. Hasanaj and Biba (2011) used a basic tagset and a large tagset containing additional features, augmenting the basic tagset. The models are experimentally evaluated, reporting a 70% accuracy on the small tagset. In a follow-up study, Salavaçi and Biba (2012) use three tagsets, a small tagset with 100 tags, a medium tagset with 150 tags, and a large tagset with 220 tags. The used corpus contains 10,000 words collected from various sources. The models are experimentally evaluated by progressively increasing the number of tokens and the number of tags, going from a small to a large tagset, in a cross-validation setting. The two models reached an average accuracy of nearly 60%.

---
[1] https://github.com/NeldaKote/Albanian-POS

The largest annotated corpus for the Albanian language is created by the Saint-Petersburg linguists' team (Arkhangelskıjet al., 2011). The Albanian National Corpus contains 16.6 million tokens and is created using the morphological parsing tool, UniParser. This corpus contains text extracted from fiction — short stories, novels, plays; non-fiction — memoirs, essays, journalism, official, religious and scientific texts. A tagset of 62 tags in total is used, including the standard tags: noun, adjective, numeral, adjective numeral, verb, adverb, pronominal clitic, preposition, conjunction, particle, preverb, prepositional article, interjection, pronoun and tags about gender, animacy, number, case, definiteness, article position, transitivity, voice, verbal representation, mood, tense, person and pronominal clitics. In the corpus, each token is given the lemma, grammatical features of the lexeme, grammatical features of the word form and the equivalent in English.

An unsupervised method using a dictionary with around 32,000 words with the corresponding POS tags of the words and a set of regular expression rules to assign POS tags to a new text is proposed by Kadriu (2013). The model is developed using the NLTK toolkit. The lemmatizer is used to convert only nouns and verbs into their lemma and is based on rules to remove or replace suffixes of the words. The authors use a tagset with 22 tags. The evaluation is done using a set of 30 random opinions, economic, and culture articles with the accuracy of 88–93%.

In the UniMorph project (Kirov et al., 2018), a small annotated morphological corpus of Albanian inflected words is extracted from Wiktionary. The annotation is done on word level without sentence context and it conforms to the UniMorph schema, where each inflected word is associated to its lemma and a set of morphological tags. The corpus contains 33,483 word forms for 589 lemmas. The corpus is used in the CoNLL-SIGMORPHON 2017 shared task (Cotterell et al., 2017) to train and test models for morphological analysis. However, as the corpus contains individual words without sentence context, the data cannot directly be used to train a part-of-speech or morphological tagger.

The corpus presented in Kabashiand and Proisl (2018) is the gold version of the corpus introduced in Kabashi and Proisl (2016). The corpus is manually annotated by two native Albanian speakers and contains 2,020 sentences, with 31,584 tokens. The authors define a full tagset as well as its mapping to Google/UD Universal POS. The full tagset contains a total of 79 tags divided into 16 main classes: noun (4 tags), verb (14 tags), adjective (5 tags), adverb (3 tags), pronoun (14 tags), preposition (1 tag), conjunction (6 tags), number (2 tags), particle (19 tags), interjection (1 tag), article (1 tag), pronominal (3 tags), abbreviation (2 tags), punctuation (2 tags), 1 tag for the non-linguistic element and 1 tag for emoticon. There are no tags for the number, gender, definiteness or case of nouns. The corpus is used to train and test six statistical POS taggers. For five taggers: the HMM-based HunPos tagger, OpenNLP tagger, TreeTagger, SoMeWeTa tagger, and Stanford POS Tagger the evaluation is done in five scenarios combining the three tagsets in training and testing phases. The models' accuracy varies from 86.96% to 95.10%. The best performing tagger, SoMeWeTa achieves the accuracy of 95.10% on the Google UPOS and 91.00% on the full tagset.

## 2.2 Lemmatization and Stemming

The first rule-based stemming algorithm for the Albanian language is developed by Karanikolas (2009). This algorithm is based on longest-match suffix removal. Also, a list of 470 stop words is defined. A corpus containing only 5,000 words is used to generate the stems. The evaluation of the algorithm is done by randomly selecting 500 words, which are evaluated manually. The results show an accuracy of 80%. In this case, it is important to emphasize that the author and the evaluators are not native Albanian speakers.

The first stemming algorithm for the Albanian language developed by a native speaker is presented by Sadiku and Biba (2012). The rule-based JStem algorithm is based on word formation with affixes. The algorithm uses 134 rules and a stopword list to remove suffixes and prefixes of a word. The rules do not cover plural formation, feminine, masculine and neuter gender formation, nor compound words. Biba and Gjati (2014) present an extension of the JStem algorithm to find stems also for compound words. The standalone performance of the algorithm is not evaluated, but downstream evaluation results indicate it to improve document classification accuracy.

## 3. The Albanian Language

The Albanian language is an independent branch of the Indo-European language family. It is an official language in Albania and Kosovo, and the official regional language in Ulcinj in Montenegro and in some municipalities of the New Republic of Macedonia. Also, Albanian is spoken in some areas in Greece, southern Serbia, in some provinces in southern Italy and by albanian communities all over the world. The Albanian language has two dialects, Gheg and Tosk. The official, contemporary language is based on the Tosk dialect and used the Roman alphabet (Hamp, 2016). Its alphabet contains 36 letters, 29 consonants where 20 consonants are single letters (*b, c, ç, d, f, g, h, j, k, l, m, n, p, q, r, s, t, v, x, z*), and 9 are combination of two letters (*dh, gj, ll, nj, rr, sh, th, xh, zh*), and 7 vowels (*a, e, ë, i, o, u, y*).

In the following, we describe the main characteristics of Albanian morphology, based on the description of Agalliu et al. (2002). The Albanian language has ten parts of speech: noun, verb, pronoun, adjective, adverb, preposition, conjunction, particle, numeral, and interjection.

There are five cases: nominative, genitive, dative, accusative and ablative, where the inflection of genitive and dative are always the same. It is the context that makes the distinction between these two cases.

### 3.1 Noun

Albanian nouns have four morphological categories as described below:
- Gender: All nouns have gender, most feminine or masculine, and a small number are neuter. Many

nouns are heterogeneous, meaning that they have different gender in singular and plural form.
- Number: Nouns have singular or plural number. In some cases, the singular and plural have the same form, e.g. "një punë" (singular, engl. job) and "disa punë" (plural, engl. jobs).
- Case: There are five cases: nominative, genitive, dative, accusative, and ablative. In genitive and dative, nouns mostly have the same form, differing only by the preceding article in genitive.
- Definiteness: There are two categories: indefinite and definite. The two are distinguished in the form: indefinite form "(një) vajzë" (engl. (one) girl), definite form "vajza" (engl. (the) girl).

### 3.2 Adjective

Adjectives have the morphological categories of gender, number, and case dependent on related nouns. In addition, adjectives have a degree of gradation. There are three grades: positive, comparative, and superlative. The gradation of an adjective is done by combining the base word with comparative particles, or adverbs. In the case of comparative gradation, the particle "më" is used, e.g. positive: "i madh" (engl. big), comparative: "më i madh" (engl. bigger) and superlative: "madheshtor" (engl. the biggest). Adjectives can be positioned before or after the noun that it describes, and they can be preceded by an article or not.

### 3.3 Numerals

Albanian numerals are classified as cardinal and ordinal numbers. Ordinal numbers have the same morphological categories as adjectives, except for gradation, and are always preceded by an article.

### 3.4 Pronoun

The pronouns are classified into 7 types according to their meaning and sense:
- Possessive Pronouns: i im, e, tu, i, saj, etc.
- Interrogative Pronouns: kush, cili, cila, etc.
- Demonstrative Pronouns: ky, kjo, ai, ajo, etc.
- Subject Pronouns: unë, ti, ai, ajo, ne, ju, ata, ato.
- Relative Pronouns: që, i cili, e cila, të cilat, etc.
- Indefinite Pronouns: dikush, askush, ndonjë, etc.
- Reflexive Pronouns: veten, vetveten.

The subjective pronouns, "ai" and "ajo" can be demonstrative pronouns in case they are followed by a noun. Some pronouns like some nouns and adjectives can be preceded by an article. There is also a short form of subjective pronouns.

### 3.5 Verb

Albanian verbs have the following morphological categories:
- Person: Verbs can be in first, second or third person.
- Number: Verb can be in singular or plural.
- Voice: There are two voices: active and non-active that includes passive, reciprocal and middle.
- Mood: There are five moods: indicative, subjunctive, optative, admirative and imperative.
- Tense: There are three tenses: present, past and future. The tense can have different aspects in different moods of the verbs.
- Finiteness: finite and non-finite that include infinitive, participle, gerundive and absolutive.

Furthermore, verbs are differentiated as auxiliary, modal, reflexive, reciprocal verbs and the participle form of the verb. Verbs have the most complex inflection of Albanian word categories.

### 3.6 Adverb

Adverbs are part of speech that do not change. There are five types of adverbs: manner, quantity, time, location and cause. Like adjectives, adverbs have the morphological category of gradation (positive, comparative, and superlative). The gradation of an adverb is created by combining the base word with comparative particles, or adverbs.

### 3.7 Preposition, Particle, Interjection

These part-of-speech do not inflect and therefore have no additional morphological categories.

### 3.8 Conjunction

There are two main categories of conjunctions in Albanian, subordinating and coordinating conjunctions. There are 14 types of conjunctions, 4 of coordinating conjunction (additive, separate, opposing and concluding) and 10 of subordinating conjunction (causative, local, temporal, causal, intentional, comparative, modal, conditional, derivative, and permissive).

## 4. Albanian Corpus and Annotation

In section 3, we discussed the morphological structure of the Albanian language. In this section, we present the corpus that we annotate, and the process based on the Universal Dependencies scheme.

### 4.1 Text Selection

The corpus contains 117,686 tokens (6,644 sentences) of written Albanian collected from different text sources: a collection of sentences from two fiction books, a collection of sentences from an Albanian grammar book to include grammatically complex sentences and morphological variation, and sentences sampled from the Albanian section of the Leipzig Corpora Collection (Goldhahn et al., 2012), which is based on web-crawled text and Wikipedia. Additionally, we include an artificially created, 66,911 token (17,042 sentences) collection of simple sentences based on a list of Albanian verbs collected from the UniMorph project (Kirov et al., 2018).

As a starting point, annotations done by Salavaçi and Biba (2012) are obtained and converted to the Universal Dependencies (UD) scheme, using a conversion script based largely on a lookup table. These annotations were subsequently manually corrected by the first author (a native Albanian speaker). Each tag of the Salavaçi and Biba (2012) large tagset was mapped to the corresponding UD part-of-speech tag and the morphological feature. As the original annotations only contain part-of speech tags, the lemma was added for each token. The annotations span 476 sentences (7374 tokens), selected from two fiction and a grammar book: Kadare (2011), Fojhtvanger (1999) and Lafe et al. (1979).

These annotations are used as a seed corpus for training a tagger using the Turku Neural Parser Pipeline, which is used to pre-tag the remainder of the corpus. This pre-tagged version was reviewed by two native speakers with a good command of the task.

The corpus is split into training, development and test sections, where sentences from the same document are always grouped into the same section to avoid unintentional lexical overlap. All artificially created short sentences based on UniMorph verbs are placed into the training section to prevent from overly optimistic evaluation results due to artificially simple sentences. The main statistics of the corpus are summarized in Table 1.

|  | Train | Devel | Test |
|---|---|---|---|
| **Tokens** | | | |
| Books, web, Wikipedia | 93,621 | 11,375 | 12,690 |
| Artificial UniMorph | 66,911 | — | — |
| Total | **160,532** | **11,375** | **12,690** |
| **Sentences** | | | |
| Books, web, Wikipedia | 5,324 | 612 | 708 |
| Artificial UniMorph | 17,042 | — | — |
| Total | **22,366** | **612** | **708** |

Table 1: Corpus statistics

### 4.2 Annotation in Universal Dependencies Scheme

For each word in the corpus, we have defined the lemma, the POS tag and the morphological features corresponding to the POS tag. In Table 2, we have specified the POS tags and morphological features used to annotate the corpus.

| Part of speech | POS tag | Features |
|---|---|---|
| Noun | NOUN | Case |
| | | Definite |
| | | Gender |
| | | Number |
| Proper noun | PROPN | |
| Adjective | ADJ | Degree |
| Numerals | NUM | NumType |
| Pronoun | PRON | Case |
| | | Gender |
| | | Number |
| | | Person |
| | | PronType |
| Verb | VERB / AUX | Mood |
| | | Number |
| | | Person |
| | | Tense |
| | | VerbForm |
| Adverb | ADV | AdvType |
| Preposition | ADP | |
| Conjunction | CCONJ / SCONJ | |
| Part | DET | |
| Interjection | INTJ | |
| Symbol | SYM | |
| Punctuation | PUNCT | |

Table 2: Part-of-speech tags and morphological features used in the annotation

Verbs present a particular challenge, since many verb tenses are formed by combining two verbs or by combining a verb preceded by an article or an adverb. These are always tagged regarding the part-of-speech and inflections of individual words, as the combined meaning is not present in any single individual word and in principle each word must receive individual tags in UD. For example, the verb "*bëj*" (engl. *do*), in present perfect tense of the indicative mood is inflected as "*ka bërë*" (engl. *have done*), and is tagged as shown in Table 3.

| Token | Lemma | POS | Features |
|---|---|---|---|
| ka | kam | AUX | Mood=Ind\|Number=Sing\|Person=3\|Tense=Pres |
| bërë | bëj | VERB | VerbForm=Part |

Table 3: Annotation example

Furthermore, the word "*kam*" can also appear as a standalone main verb in which case it is naturally tagged as VERB rather than an auxiliary. The same applies also to the simple future tense, which is created by combining particles "*do*" and "*të*" with an inflected verb. As none of these words individually hold the future tense, "*do*" and "*të*" are tagged as particles and the verb is tagged depending on the actual case of the inflected word (present or past). Therefore, the future tense is not explicitly annotated in the corpus. The same holds also for the passive voice, which except for the present and imperfect tense, is formed by combining the article "*u*" with the verb in the active or using the "*jam*" (engl. *to be*) verb instead of "*kam*" (engl. *to have*) in the composite tense. For example, the passive form of the verb "*ka bërë*" (engl. have done; active, present perfect tense, indicative mood) is "*jam bërë*".

Similarly, in adjectives the comparative and superlative gradations are formed by combining a comparative or superlative particle with a positive adjective. In such cases, the gradation is not marked. This follows the style used in the UD English-EWT corpus, where for example "*more careful*" is tagged as an adverb and a positive form of the adjective. Therefore, only in the case of positive adjectives the *Degree=Pos* feature is used in the Albanian corpus.

Furthermore, all foreign words, mostly nouns and proper names, are identified with the *Foreign=Yes* feature. The annotation of proper names (PROPN) include person and animal names, as well as names of institutions, movie titles, etc.

## 5. Turku Neural Parser Pipeline

The system used in our experiments is the Turku Neural Parser Pipeline (Kanerva et al., 2018), a full parser pipeline for end-to-end analysis from raw text into UD. The pipeline includes sentence and word segmentation, part-of-speech and morphological tagging, syntactic parsing, and lemmatization. All these components are wrapped into a single system, and they directly support training with CoNNL-U formatted corpora. The Turku Pipeline was ranked second on the LAS and MLAS metric, and first on the BLEX metric of the CoNLL-2018 Shared Task, making it highly competitive. Further, as the MLAS and BLEX metrics explicitly include the accuracy of morphological analysis and lemmatization, the parsing pipeline is especially relevant to the present work.

The text segmentation component in the Turku pipeline is based on UDPipe (Straka and Straková, 2017), where the token and sentence boundaries are jointly predicted using a single-layer bidirectional GRU network. Universal part-of-speech tag (UPOS) and morphological features (FEATS) are predicted with a modified version of the one published by Dozat et al. (2017), a time-distributed classifier over tokens in a sentence embedded using a bidirectional LSTM network. The tagger has two separate classification layers, one for universal part-of-speech and one for morphological features. The bidirectional encoding is shared between both classifiers. The lemmatizer component by Kanerva et al. (2019) is a sequence-to-sequence model, where the lemma is generated one character at a time from the given input word form and morphological features.

The models are trained with default parameter settings using Albanian word embeddings created by Bojanowski et al. (2017) using the fastText tool on Wikipedia data. During training, the development set is used for early stopping and the held-out test section is preserved for evaluation.

## 6. Experimental Evaluation

The tagger system described above is trained and evaluated on the data described in Section 4. The performance on the test section is reported in Table 3, measured using the official evaluation script[2] from the CoNLL 2018 shared task on multilingual parsing from raw text to Universal Dependencies (Zeman et al., 2018). We measure accuracy on token and sentence segmentation, part-of-speech tagging, morphological features and lemmatization. We deem that the performance of each step of the pipeline is at a sufficiently high level in order to support many downstream NLP applications in practice.

| Target | Accuracy [%] |
| --- | --- |
| Token | 99.88 |
| Sentence | 99.51 |
| POS | 92.74 |
| Features | 85.31 |
| Lemmas | 89.95 |

Table 3: Evaluation results on the tasks of text segmentation, POS and morphological tagging, and lemmatization.

## 7. Conclusions and Future Work

We have created a morphologically analyzed corpus of the Albanian language comprising 118,000 tokens, with an additional 67,000 tokens of short, artificial examples only present in the training data. The annotation is fully manually reviewed and conforms to the UD morphological tagsets. On this corpus, we have trained a competitive end-to-end text segmentation and tagging pipeline, achieving promising performance in terms of tagging and lemmatization accuracy. We expect the Albanian POS and morphological tagger, as well as the lemmatizer, to be able to serve as practical building blocks in many language processing applications for Albanian.

---

[2] https://universaldependencies.org/conll18/conll18_ud_eval.py

Both the corpus and the trained model for the Turku Neural Parser Pipeline are available online, under an open license. Presently, these comprise the only such publicly available resources for Albanian.

Since the morphological annotation conforms to the UD guidelines, this work also lays the foundation for a possible future extension with syntactic annotation and inclusion into the Universal Dependencies treebank collection. This is, however, left as a future work.

## Acknowledgements

We greatfully acknowledge the support of the EU Erasmus+ KA107 Internation Credit Mobility Programme and the partners at the Polytechnic University of Tirana and Åbo Akademi University, as well as Academy of Finland, CSC – IT Center for Science, and the NVIDIA Corporation GPU Grant Program.

## 8. Bibliographical References

Agalliu F., Angoni E., Demiraj Sh., et al. (2002). Gramatika e gjuhës shqipe 1 /Morfologjia. Instituti i Gjuhësisë dhe i Letërsisë (Akademia e Shkencave e RSH), 1.

Arkhangelskıj, T., Danıel, M., Morozova, M. and Rusakov, A. (2011). Korpusı i Gjuhës Shqıpe: Drejtımet Kryesore të Punës". Proceedings of the conference on Albanıan And Balkan Languages, pp. 635-683.

Biba, M. and Gjati, E. (2014). Boosting Text Classification through Stemming of Composite Words. Recent Advances in Intelligent Informatics, Advances in Intelligent Systems and Computing, 235:185-194.

Bojanowski, P., Grave, E., Joulin, A. and Mikolov, T. (2017). Enriching Word Vectors with Subword Information. Transactions of the Association for Computational Linguistics.

Cotterell, R., Kirov, C., Sylak-Glassman, J., Walther, G., Vylomova, E., Xia, P., Faruqui, M., Kübler, S., Yarowsky, D., Eisner, J. and Hulden, M. (2017). CoNLL-SIGMORPHON 2017 Shared Task: Universal Morphological Reinflection in 52 Languages. Proceedings of the CoNLL SIGMORPHON 2017 Shared Task: Universal Morphological Reinflection.

Dozat, T., Qi, P. and Manning, C.D. (2017). Stanfords graph-based neural dependency parser at the CoNLL 2017 shared task. In Proceedings of the CoNLL 2017 Shared Task: Multilingual Parsing from Raw Text to Universal Dependencies.

Fojhtvanger, L. (1999). Çifutka e Toledos. Shtëpia botuese Dituria. ISBN 99927-31-36-2

Goldhahn, D., Eckart, T. and Quasthoff, U. (2012). Building Large Monolingual Dictionaries at the Leipzig Corpora Collection: From 100 to 200 Languages. In Proceedings of the 8th International Conference on Language Resources and Evaluation (LREC'12).

Hamp, E.P. (2016). Albanian language, Encyclopedia Britannica.

Hasanaj B. and Biba M. (2011). A Part of Speech Tagging Model for Albanian: An innovative solution. LAP Lambert Academic Publishing. ISBN 13: 9783659223273. Master's Thesis.

Kabashi, B. and Proisl, Th. (2018). Albanian Part-of-Speech Tagging: Gold Standard and Evaluation. Proceedings of the Eleventh International Conference on Language Resources and Evaluation (LREC'18), pp. 2593–2599.

Kabashi, B. and Proisl, Th. (2016). A Proposal for a Part-of-Speech Tagset for the Albanian Language. Proceedings of the Tenth International Conference on Language Resources and Evaluation (LREC'16), pp. 4305–4310.

Kadare, H. (2011). Kohë e pamjaftueshme. Shtëpia botuese Onufri. ISBN 978-999556-87-51-9

Kadriu, A. (2013). NLTK Tagger for Albanian using Iterative Approach. Proceedings of the 35th International Conference on Information Technology Interfaces, pp. 283-288.

Kanerva, J., Ginter, F., Miekka, N., Leino, A., and Salakoski, T. (2018). Turku Neural Parser Pipeline: An end-to-end system for the CoNLL 2018 shared task. In Proceedings of the CoNLL 2018 Shared Task: Multilingual Parsing from Raw Text to Universal Dependencies.

Kanerva, J., Ginter, F. and Salakoski, T. (2019). Universal Lemmatizer: A sequence to sequence model for lemmatizing Universal Dependencies treebanks. arXiv preprint arXiv:1902.00972.

Karanikolas, N. (2009). Bootstrapping the Albanian Information Retrieval. Fourth Balkan Conference in Informatics, pp. 231-235.

Kirov, Ch., Cotterell, R., Sylak-Glassman, J., Walther, G., Vylomova, E., Xia, P., Faruqui, M., Mielke, S., McCarthy, A., Kubler, S., Yarowsky, D., Eisner, J. and Hulden, M. (2018). UniMorph 2.0: Universal Morphology. In Proceedings of the Eleventh International Conference on Language Resources and Evaluation (LREC'18).

Lafe, V., Buxheli, L., and Basha, N. (1979). Libri i gjuhës shqipe 1. Shtëpia botuese e librit shkollor. Tiranë.

Nivre, J., De Marneffe, M.C., Ginter, F., Goldberg, Y., Hajic, J., Manning, C.D., McDonald, R., Petrov, S., Pyysalo, S., Silveira, N., Tsarfaty, R. and Zeman, D. (2016). Universal Dependencies v1: A multilingual treebank collection. In Proceedings of the Tenth International Conference on Language Resources and Evaluation (LREC'16).

Piton, O., Lagji, K. and Përnaska, R. (2007). Electronic Dictionaries and Transducers for Automatic Processing of the Albanian Language. 12th International Conference on Applications of Natural Language to Information Systems, 4592:407-413.

Piton, O. and Lagji, K. (2007). Morphological study of Albanian words, and processing with NooJ. In Proceedings of the 2007 International NooJ Conference, pp. 189-205.

Sadiku, J. and Biba, M. (2012). Automatic Stemming of Albanian Through a Rule-based Approach. Journal of International Scientific Publications: Language, Individual Society, Volume 6.

Salavaçi, E. and Biba M. (2012) Enhancing Part-of-Speech Tagging in Albanian with Large Tagsets. Master's Thesis.

Straka, M. and Straková, J. (2017). Tokenizing, POS tagging, lemmatizing and parsing UD 2.0 with UDPipe. In Proceedings of the CoNLL 2017 SharedTask: Multilingual Parsing from Raw Text to Universal Dependencies.

Trommer, J. and Kallulli, D. (2004). A Morphological Tagger for Standard Albanian. In Proceedings of the Fourth International Conference on Language Resources and Evaluation (LREC'04).

Zeman, D., Hajič, J., Popel, M., Potthast, M., Straka, M., Ginter, F., Nivre, J. and Petrov, S. (2018). CoNLL 2018 Shared Task: Multilingual Parsing from Raw Text to Universal Dependencies. In Proceedings of the CoNLL 2018 Shared Task: Multilingual Parsing from Raw Text to Universal Dependencies.